\documentclass[conference]{IEEEtran}
\IEEEoverridecommandlockouts
\usepackage{amsmath,amssymb,amsfonts}
\usepackage{algorithmic}
\usepackage{graphicx}
\usepackage{textcomp}
\usepackage{xcolor}
\usepackage{amsfonts}
\usepackage{booktabs}  
\usepackage{colortbl}
\usepackage{marvosym}%
\usepackage{multirow}
\usepackage{makecell}
\usepackage{pifont} 
\usepackage{arydshln}
\usepackage{bigstrut}
\usepackage[hyperindex, plainpages=false, colorlinks=true, linkcolor=blue, citecolor=blue, urlcolor=black]{hyperref}
\usepackage{subcaption}
\def\BibTeX{{\rm B\kern-.05em{\sc i\kern-.025em b}\kern-.08em
    T\kern-.1667em\lower.7ex\hbox{E}\kern-.125emX}}
\begin{document}

\title{Revisiting DETR for Small Object Detection via Noise-Resilient Query Optimization}

\author{
    \IEEEauthorblockN{Xiaocheng Fang\textsuperscript{\rm1*}, Jieyi Cai\textsuperscript{\rm1*}, Huanyu Liu\textsuperscript{\rm1}, Wenxiu Cai\textsuperscript{\rm1}, Yishu Liu\textsuperscript{\rm2\Letter}, Bingzhi Chen\textsuperscript{\rm3,1\Letter}}
    \IEEEauthorblockA{\textsuperscript{\rm1}South China Normal University, Guangzhou, China}
    \IEEEauthorblockA{\textsuperscript{\rm2}Harbin Institute of Technology, Shenzhen, China}
    \IEEEauthorblockA{\textsuperscript{\rm3}Beijing Institute of Technology, Zhuhai, China}
    \IEEEauthorblockA{\tt\small \{fangxiaocheng,caijieyi,20222005228,101088\}@m.scnu.edu.cn,\\ {\tt\small liuyishu.smile@gmail.com}, {\tt\small chenbingzhi@bit.edu.cn}}\thanks{$^{*}$ Equal contribution.}\thanks{\Letter Corresponding authors: Bingzhi Chen and Yishu Liu.}\thanks{This work was supported in part by the Guangdong Basic and Applied Basic Research Foundation (No. 2025A1515010225), in part by the National Natural Science Foundation of China (No. 62302172), and in part by the Shenzhen Fundamental Research Fund (No. JCYJ20240813105900002).}
}

\maketitle

\begin{abstract}
Despite advancements in Transformer-based detectors for small object detection (SOD), recent studies show that these detectors still face challenges due to inherent noise sensitivity in feature pyramid networks (FPN) and diminished query quality in existing label assignment strategies. In this paper, we propose a novel Noise-Resilient Query Optimization (NRQO) paradigm, which innovatively incorporates the Noise-Tolerance Feature Pyramid Network (NT-FPN) and the Pairwise-Similarity Region Proposal Network (PS-RPN). Specifically, NT-FPN mitigates noise during feature fusion in FPN by preserving spatial and semantic information integrity. Unlike existing label assignment strategies, PS-RPN generates a sufficient number of high-quality positive queries by enhancing anchor-ground truth matching through position and shape similarities, without the need for additional hyperparameters. Extensive experiments on multiple benchmarks consistently demonstrate the superiority of NRQO over state-of-the-art baselines.
\end{abstract}


\section{Introduction}
As a fundamental task in computer vision, object detection entails both localizing and classifying objects. Existing methods primarily fall into two categories: CNN-based and Transformer-based detectors. CNN-based detectors, such as Faster R-CNN~\cite{ren2015faster}, RetinaNet~\cite{lin2017focal}, and FCOS~\cite{tian2019fcos}, rely on manually designed components, including anchors, ROI thresholds, and NMS, resulting in a complex, hyperparameter-intensive pipeline that limits end-to-end optimization. In contrast, Transformer-based detectors, such as DETR~\cite{carion2020end}, Conditional-DETR~\cite{meng2021conditional}, and Salience-DETR~\cite{hou2024salience}, simplify the process by leveraging self-attention, eliminating the need for hand-crafted components, and enhancing global context modeling. Despite these advancements, directly applying general object detectors to small object detection tasks leads to a substantial drop in accuracy.

Small object detection (SOD) remains a significant challenge in computer vision. Compared to general-scale objects, small objects present unique difficulties due to their limited size, low resolution, and lack of distinctive features, which make them harder to localize and classify accurately~\cite{noh2019better,yang2022querydet}. Furthermore, small objects are more susceptible to noise and occlusion, further complicating detection tasks. Despite these challenges, SOD holds substantial significance in both research and practical applications, such as autonomous driving~\cite{hu2023planning}, aerial surveillance~\cite{fang2023differentiated}, and medical imaging~\cite{chen2024cariesxrays}.

\begin{figure}[t]
    \centering
    \begin{subfigure}{0.49\linewidth}
        \centering
        \includegraphics[width=\linewidth]{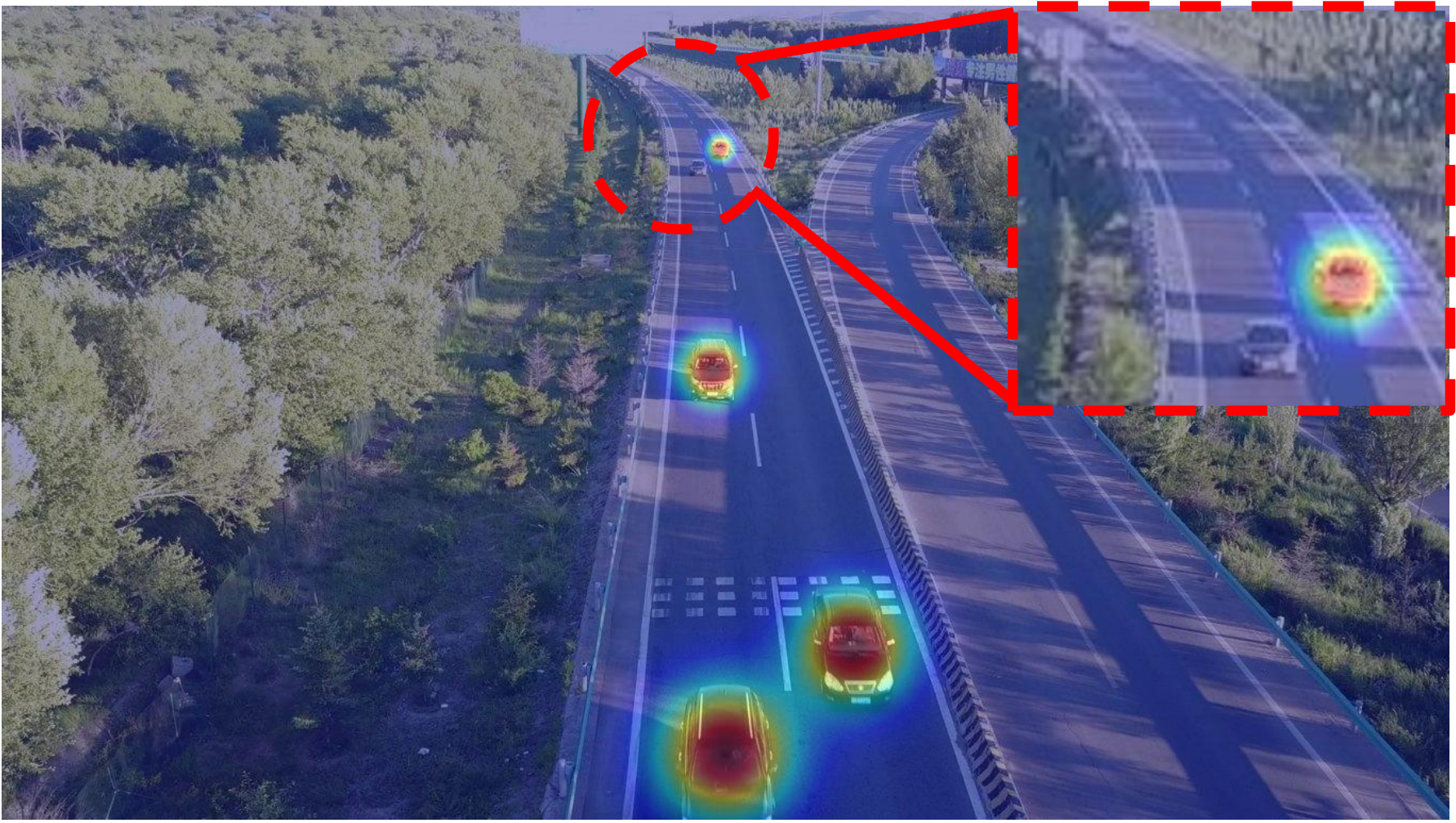}
        \caption{FPN-based Heatmap}
        \label{Fig_1a}
    \end{subfigure}
    \hfill
    \begin{subfigure}{0.49\linewidth}
        \centering
        \includegraphics[width=\linewidth]{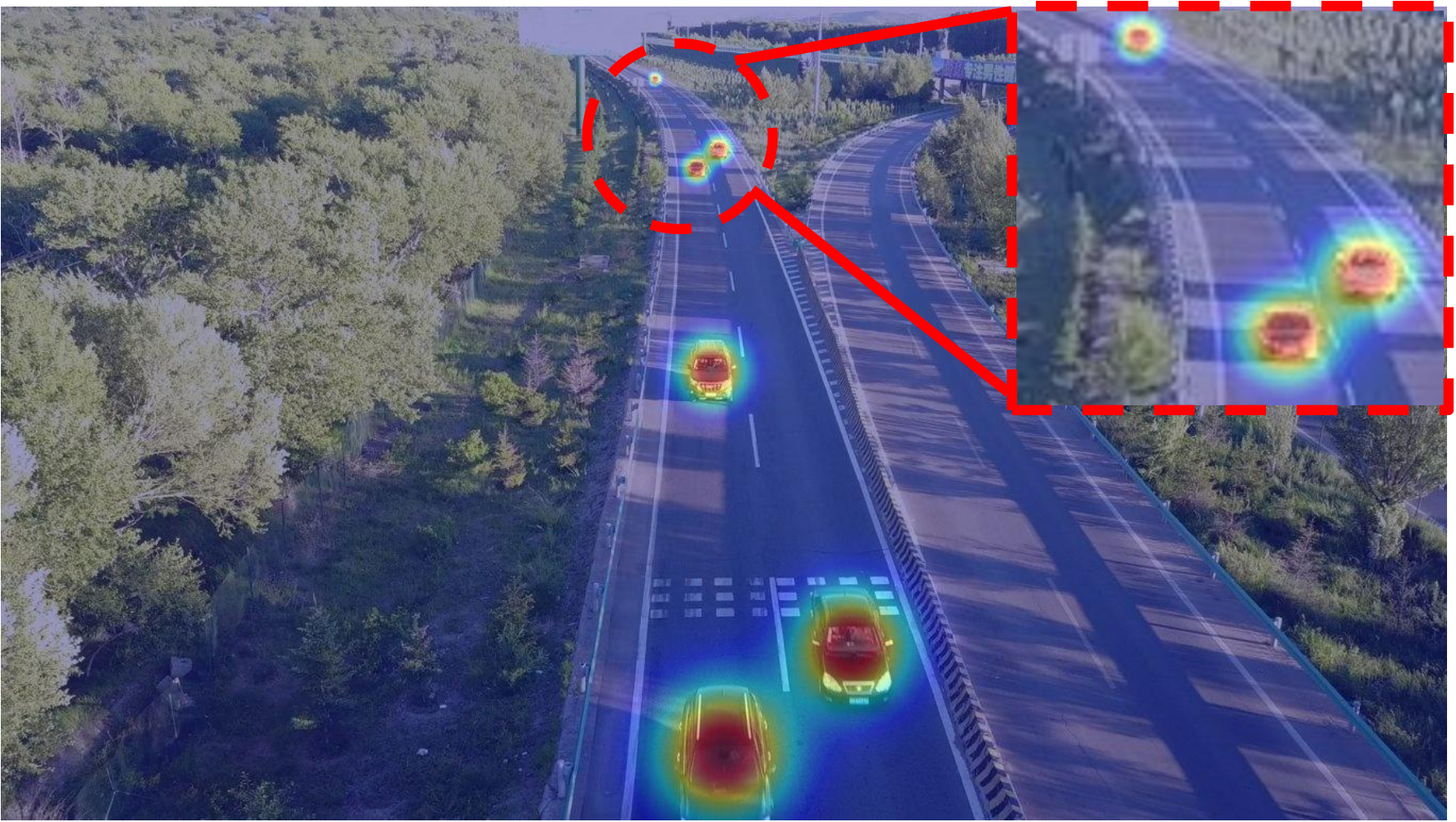}
        \caption{Our NT-FPN Heatmap}
        \label{Fig_1b}
    \end{subfigure}
    
    \vspace{0.1cm}  
    
    \begin{subfigure}{0.49\linewidth}
        \centering
        \includegraphics[width=\linewidth]{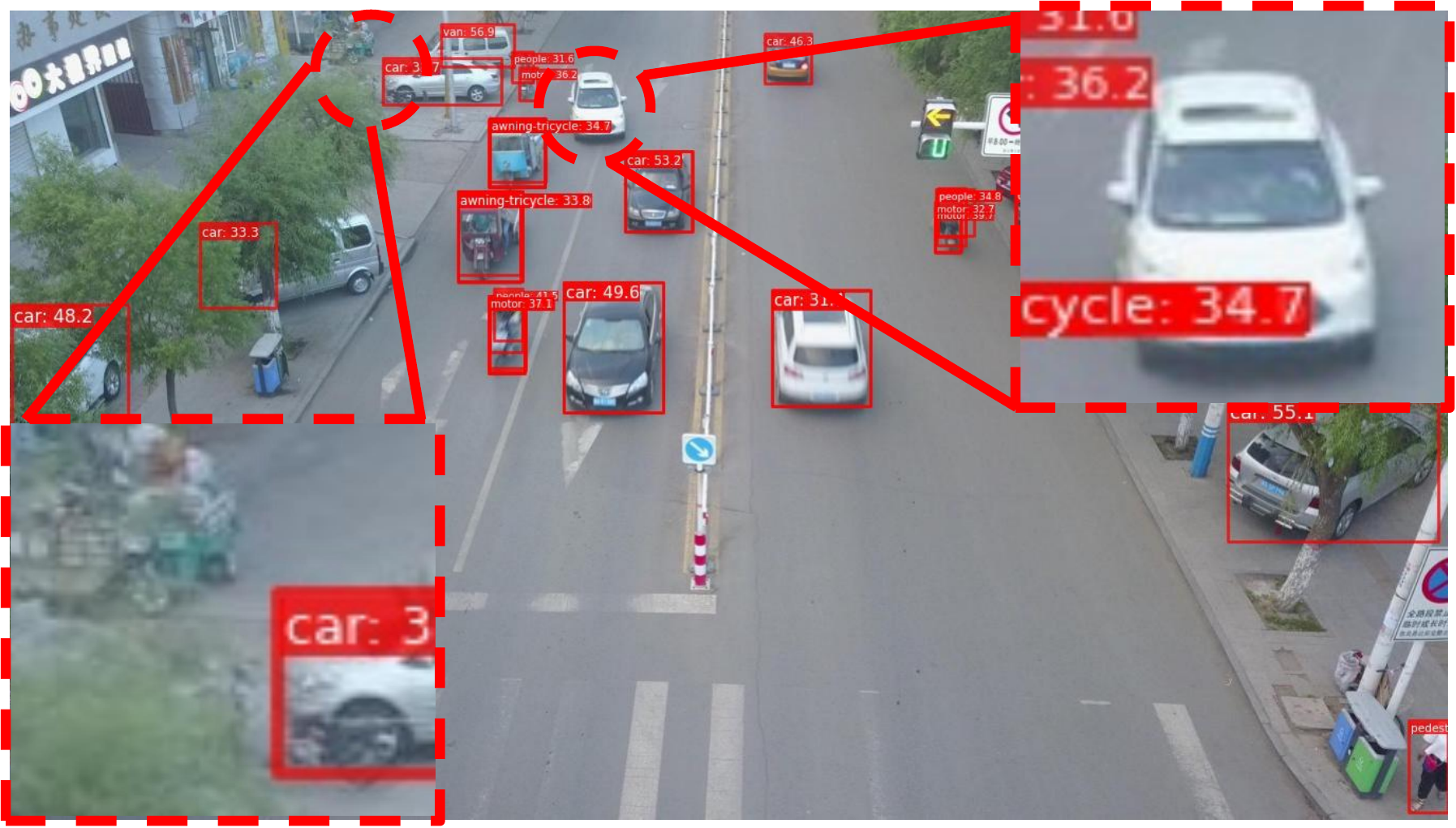}
        \caption{DotD-based Result}
        \label{Fig_1c}
    \end{subfigure}
    \hfill
    \begin{subfigure}{0.49\linewidth}
        \centering
        \includegraphics[width=\linewidth]{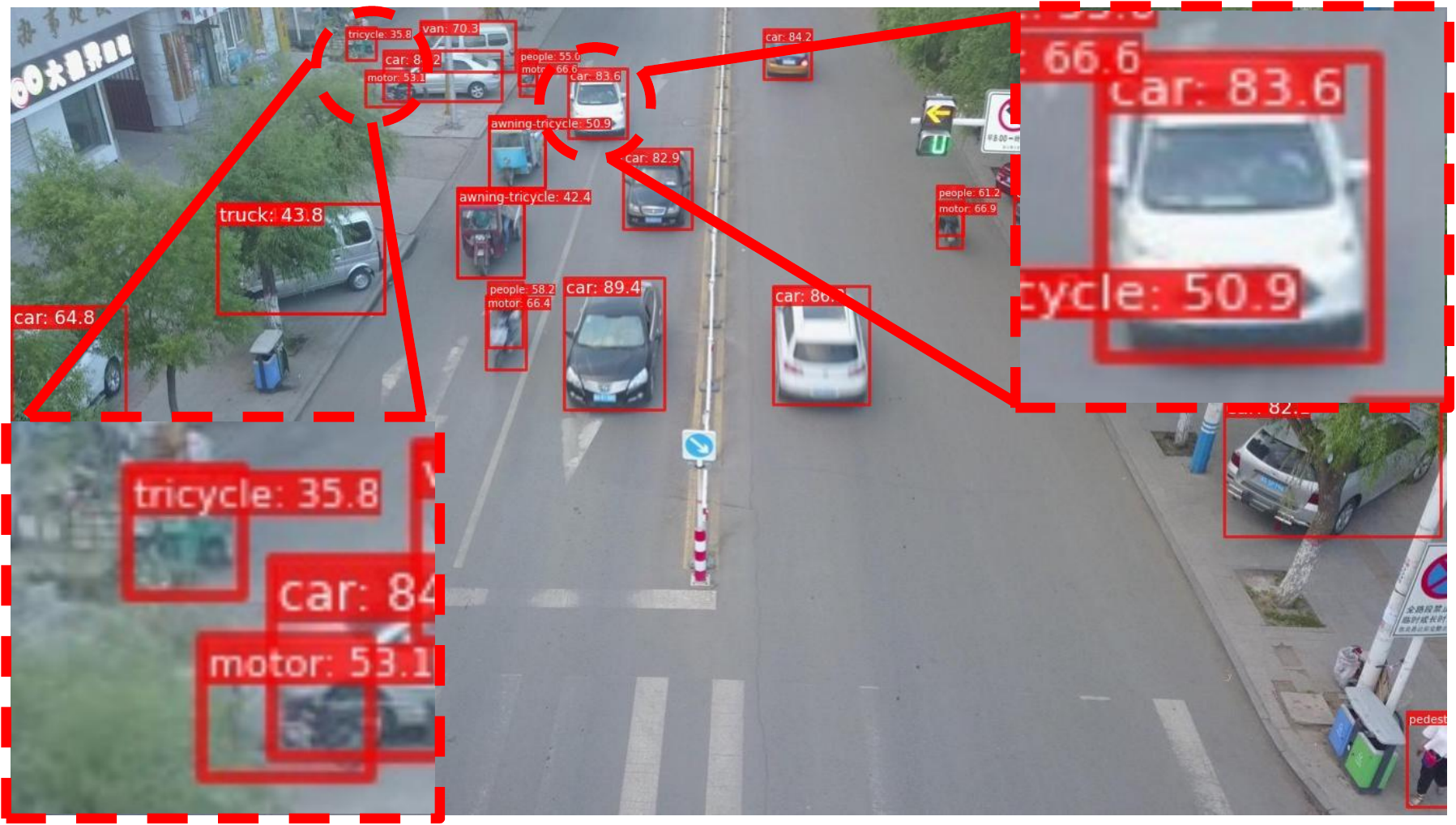}
        \caption{Our PS-RPN Result}
        \label{Fig_1d}
    \end{subfigure}
    \caption{Comparison of query heatmaps and detection results between traditional methods and our methods.}
    \vspace{-0.2in}
    \label{Fig_1}
\end{figure}

\begin{figure*}[t]
\centering
\includegraphics[width=0.99\linewidth]{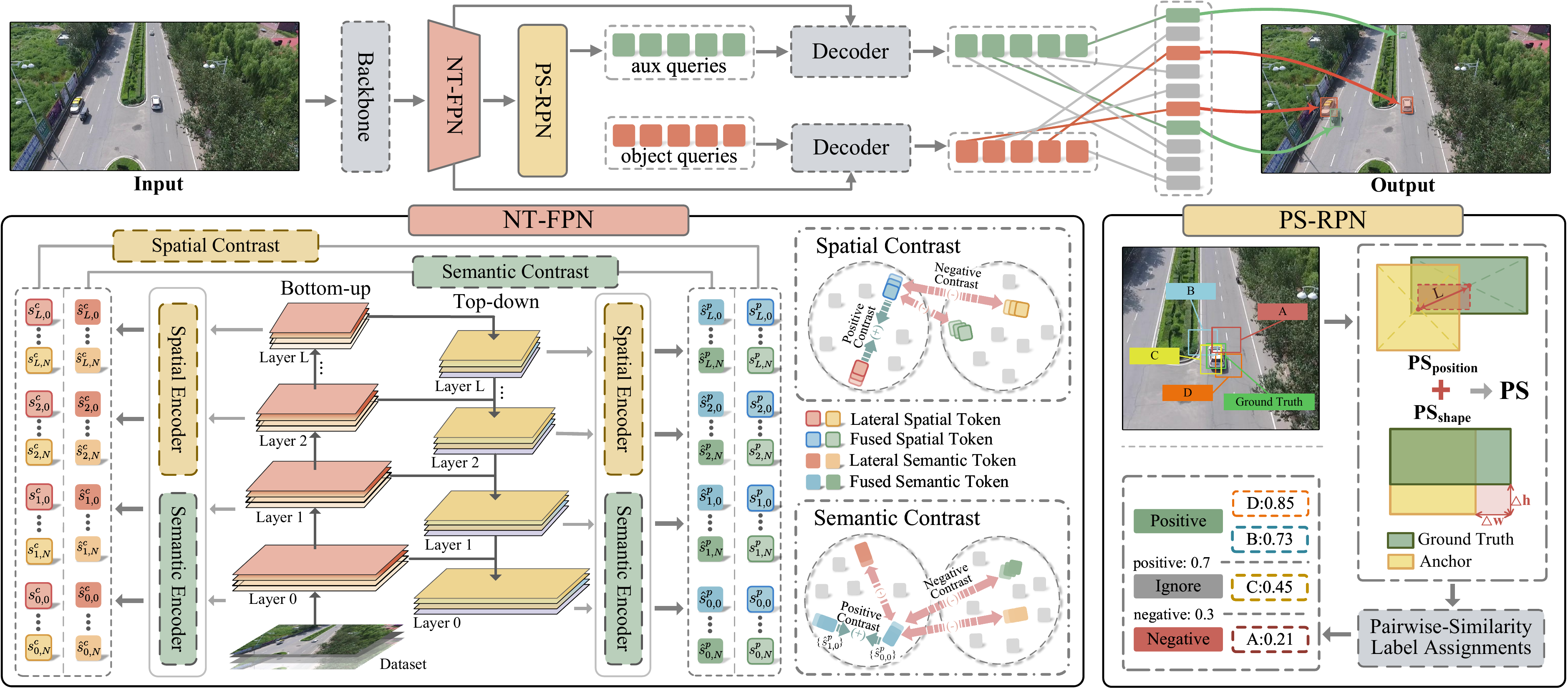}
\caption{Illustration of our proposed NRQO paradigm, which consists of two key modules: (1) NT-FPN, reducing noise in FPN and enhancing small object representation by preserving spatial and semantic integrity; (2) PS-RPN, improving the quality of positive queries for small objects by modeling position and shape similarities between anchors and ground truth.}
\vspace{-0.2in}
\label{Fig_3}
\end{figure*}

In small object detection, the feature pyramid network (FPN)~\cite{lin2017feature} has become a cornerstone strategy due to its ability to integrate multi-scale features, thereby improving the detection of fine-grained details and global context for small target objects. Building on this foundation, numerous studies have focused on optimizing FPN-based networks. For example, Gong et al.~\cite{gong2021effective} introduce a fusion factor to adjust the influence of high-level features, mitigating the attenuation of small object semantics during feature fusion. Additionally, Gao et al.~\cite{gao2023global} enhance high-level semantic features by rotating them at four different angles, concatenating them along the channel dimension, and passing them through convolution layers to improve interaction across various perspectives, thus reinforcing global semantic information. Hu et al.~\cite{hu2022towards} utilize adaptive hierarchical upsampling to generate FPN features, compensating for low-level distortions and addressing the dilution issue caused by FPN fusion.

Label assignment strategies play a crucial role in small object detection. These strategies are typically categorized into hard-label and soft-label approaches, depending on whether labels are strictly classified as negative or positive. Traditional intersection-over-union (IoU) metrics in SOD are highly sensitive to small positional deviations between predicted and ground truth boxes~\cite{dong2023control}. To address this, DotD~\cite{xu2021dot} introduces a metric based on the relative distance between object centers, considering the average object size in the dataset. NWD~\cite{wang2021normalized} further refines this approach by modeling bounding boxes as 2D Gaussian distributions and applying normalized Wasserstein distance to minimize the impact of small location deviations. Building on NWD, NWD-RKA~\cite{xu2022detecting} mitigates excessive negative labels by sampling the top-K highest-quality regions instead of relying on a fixed threshold.

Despite their potential to improve small object detection, most existing FPNs and label assignment strategies face two major challenges, i.e., \textbf{inherent noise sensitivity} and \textbf{diminished query quality}, both of which compromise detection precision. On the one hand, previous studies~\cite{liu2020ipg,liu2023tripartite} have demonstrated that upsampling upper-level features can introduce redundancy, while channel reduction in lateral features often leads to information loss, commonly referred to as “noise.” This noise is particularly detrimental to small object detection, as low-resolution features are especially sensitive to it. Comparing Fig.~\ref{Fig_1a} and Fig.~\ref{Fig_1b}, we observe that some small targets are missed due to noise introduced during the feature fusion process in the FPN. On the other hand, existing label assignment strategies primarily focus on bounding box positions while overlooking shape information, thereby reducing the quality of positive queries~\cite{xu2021dot,wang2021normalized}. As a result, when the detector encounters small objects with similar positions but different shapes, it struggles to assign high-quality positive queries. As shown in Fig.~\ref{Fig_1c} and Fig.~\ref{Fig_1d}, DotD-based detectors, which only consider bounding box positions, exhibit suboptimal performance in small object detection.

To address these challenges, this paper proposes a novel Noise-Resilient Query Optimization (NRQO) paradigm that incorporates the Noise-Tolerance Feature Pyramid Network (NT-FPN) and the Pairwise-Similarity Region Proposal Network (PS-RPN) to enhance small object detection. Specifically, to mitigate the detrimental effects of noise during feature fusion in FPN, the proposed NT-FPN is designed to maintain spatial and semantic integrity across multiple scales. By employing this noise-resilient strategy, NT-FPN significantly enhances small object detection. Additionally, to improve the quality of positive queries, we incorporate the PS-RPN, which refines anchor-ground truth similarity by modeling both position and shape. This dual consideration of position and shape ensures the generation of high-quality positive queries, thereby boosting small object detection performance. Our main contributions are summarized as follows:
\begin{itemize}
    \item We propose a novel Noise-Resilient Query Optimization (NRQO) paradigm to enhance small object detection by mitigating noise sensitivity and improving positive query quality through the integration of NT-FPN and PS-RPN.
    \item The proposed NT-FPN mitigates noise by preserving spatial and semantic information integrity, while PS-RPN improves positive query quality by utilizing position and shape similarities, without requiring hyperparameters.
    \item The effectiveness and superiority of our NRQO paradigm are thoroughly evaluated across multiple large-scale datasets, demonstrating its promising performance compared to state-of-the-art approaches.
\end{itemize}

\section{Methodology}
\subsection{Overview of NRQO}
The primary objective of the proposed NRQO paradigm is to reduce noise during feature fusion in the FPN and enhance the quality of positive queries for small objects. Fig. \ref{Fig_3} illustrates the pipeline of the proposed NRQO paradigm, which consists of two key modules: 1) the Noise-Tolerance Feature Pyramid Network (NT-FPN), and 2) the Pairwise-Similarity Region Proposal Network (PS-RPN).

\vspace{-0.05in}
\subsection{Noise-Tolerance Feature Pyramid Network}
To effectively detect small objects, it is crucial to address the noise introduced during multi-scale feature fusion in the FPN. While the FPN integrates features at different scales, the upsampling and channel reduction processes often introduce noise, thereby degrading small object detection performance. To mitigate this, we propose the Noise-Tolerance Feature Pyramid Network (NT-FPN), designed to reduce noise by preserving spatial and semantic integrity during feature fusion. NT-FPN prevents feature degradation and improves the quality of small object representations.

Specifically, we project the lateral feature $C_{i,j}$ into spatial and semantic representations $\{s_{i,j}^c, \hat{s}_{i,j}^c \in \mathbb{R}^{256}\}$ using spatial and semantic encoders, where $i$ denotes the FPN level and $j$ the sample index in the minibatch. Similarly, the upper-level feature $P_{i,j}$ is projected into $\{s_{i,j}^p, \hat{s}_{i,j}^p \in \mathbb{R}^{256}\}$. Our objective is to minimize the distance between the spatial information of the fused and lateral features while aligning the semantic information of the fused features with the upper-level features.

\subsubsection{Spatial Contrast} 
The spatial representations of \( C_{i,j} \) and \( P_{i,j} \) should ideally align, but channel reduction degrades \( P_{i,j} \), causing inconsistencies with \( C_{i,j} \). In our approach, given $L$ levels of an FPN and $N$ images within a minibatch, 
we define $q^s$ (\( s_{x,y}^p \)) and $k_+^s$ (\( s_{x,y}^c \)) as a positive pair for the \( x \)-th level of the top-down path and the \( y \)-th image, as they should be spatially consistent. Spatial representations from other levels or images within the minibatch, containing distinct spatial information, are treated as negative samples. We denote this set of negative samples as \( S^{-} \),
\begin{equation}
\label{Eq_5}
S^- = \left\{s^c_{i,j}, s^p_{i,j} \mid i = 0, \ldots, L, j = 0, \ldots, N; j \neq y\right\}.
\end{equation}

Thus, the spatial loss at the $x$-th level for the $y$-th image, denoted as $\mathcal{L}_s\left(q^s\right)$, is defined as follows,
\begin{equation}
\label{Eq_6}
\mathcal{L}_s\left(q^s\right) =-\log \left( 
    \frac{
        \exp \left(q^s \cdot k_+^s / \tau \right)
    }{
        \exp \left(q^s \cdot k_+^s / \tau  \right) + 
        \sum\limits_{s \in S^-} 
        \exp \left(q^s \cdot s / {\tau} \right)
    }
\right),
\end{equation}
where the temperature $\tau$ controls the concentration of features in the representation space. 

The overall spatial loss is computed as follows,
\begin{equation}
\label{Eq_7}
\mathcal{L}_s = \frac{1}{L \times N} \sum_{x=0}^{L} \sum_{y=0}^{N} \mathcal{L}_s\left(q^s\right).
\end{equation}

\subsubsection{Semantic Contrast}
Theoretically, \( P_{i+1,j} \) and \( P_{i,j} \) should have identical semantic representations. However, upsampling may introduce noise, causing semantic inconsistency. For the \( x \)-th level of the top-down path and the \( y \)-th image, we define \( q^{\hat{s}} \) (\(\hat{s}_{x,y}^p\)) and \( k_+^{\hat{s}} \) (\(\hat{s}_{x+1,y}^p\)) as a positive pair. Semantic representations from other images in the minibatch are treated as negative samples, denoted by \( \hat{S}^{-} \),
\begin{equation}
\label{Eq_8}
\hat{S}^- = \left\{\hat{s}^c_{i,j}, \hat{s}^p_{i,j} \mid i = 0, \ldots, L, j = 0, \ldots, N; j \neq y\right\}.
\end{equation}

Similarly, the semantic loss at the $x$-th level for the $y$-th image denoted as \( \mathcal{L}_{\hat{s}} \left( q^{\hat{s}} \right) \), is defined as follows,
\begin{equation}
\label{Eq_9}
\mathcal{L}_{\hat{s}}\left(q^{\hat{s}}\right) =
-\log \left( 
    \frac{
        \exp \left(q^{\hat{s}} \cdot k_+^{\hat{s}} / \tau \right)
    }{
        \exp \left(q^{\hat{s}} \cdot k_+^{\hat{s}} / \tau  \right) + 
        \sum\limits_{\hat{s} \in \hat{S}^-} 
        \exp \left(q^{\hat{s}} \cdot \hat{s} / {\tau} \right)
    }
\right).
\end{equation}

The overall semantic loss is calculated as follows,
\begin{equation}
\label{Eq_10}
\mathcal{L}_{\hat{s}} = \frac{1}{(L-1) \times N} \sum_{x=0}^{L-1} \sum_{y=0}^{N} \mathcal{L}_{\hat{s}}\left(q^{\hat{s}}\right).
\end{equation}

\begin{table*}[t]
\centering
\renewcommand\tabcolsep{2.5pt}
\renewcommand\arraystretch{1.0}
\caption{Comparison with detection approaches on the VisDrone 2019 and SODA-D.}
\vspace{-0.05in}
\resizebox{\textwidth}{!}{%
\begin{tabular}{l|c|c|cccccc|ccccccc}
\toprule
\multirow{2}{*}{\textbf{Methods}} & \multirow{2}{*}{\textbf{Ref}} & \multirow{2}{*}{\textbf{Params}} & \multicolumn{6}{c|}{\textbf{VisDrone 2019}} & \multicolumn{7}{c}{\textbf{SODA-D}} \\ 
 & & & \textbf{$\text{AP}$} & \textbf{$\text{AP}_{50}$} & \textbf{$\text{AP}_{75}$} & \textbf{$\text{AP}_{S}$} & \textbf{$\text{AP}_{M}$} & \textbf{$\text{AP}_{L}$} & \textbf{$\text{AP}$} & \textbf{$\text{AP}_{50}$} & \textbf{$\text{AP}_{75}$} & \textbf{$\text{AP}_{eS}$} & \textbf{$\text{AP}_{rS}$} & \textbf{$\text{AP}_{gS}$} & \textbf{$\text{AP}_{N}$}\\ 
\hline
\textbf{CNN-based} \bigstrut[t]\\
\hline
Faster RCNN\cite{ren2015faster} & NIPS'15 & 41M & 20.1 & 33.4 & 21.1 & 12.1 & 30.4 & 40.0 &  28.9 & 59.4 & 24.1 & 13.8 & 25.7 & 34.5 & 43.0\\
FCOS\cite{tian2019fcos} & ICCV'19 & 32M & 15.2 & 25.5 & 15.7 & 7.8 & 23.7 & 29.7 & 23.9 & 49.5 & 19.9 & 6.9 & 19.4 & 30.9 & 40.9 \\
ATSS\cite{zhang2020bridging} & CVPR'20 & 32M & 18.7 & 30.7 & 19.2 & 10.9 & 29.0 & 32.5 & 26.8 & 55.6 & 22.1 & 11.7 & 23.9 & 32.2 & 41.3 \\
YOLOX\cite{ge2021yolox} & CVPR'21 & 99M & 23.5 & 39.5 & 23.8 & 14.8 & 34.2 & 39.0 & 26.7 & 53.4 & 23.0 & 13.6 & 25.1 & 30.9 & 30.4 \\
Sparse RCNN\cite{sun2021sparse} & CVPR'21 & 106M & 8.5 & 15.5 & 8.1 & 4.9 & 12.5 & 18.6 & 24.2 & 50.3 & 20.3 & 8.8 & 20.4 & 30.2 & 39.4  \\    RFLA~\cite{xu2022rfla}& ECCV'22& 45M & 25.4 &42.2 &26.0 &17.9 &34.4 &45.7 & 29.7 & 60.2 & 25.2 & 13.2 & 26.9 & 35.4 & 44.6\\
CFINet~\cite{yuan2023small}& ICCV'23 &49M&26.0 &45.3 &26.1 &18.3 &35.3 &49.9 & 30.7 & 60.8 & 26.7 & 14.7 & 27.8 & 36.4 & 44.6\\
\hline
\textbf{Transformer-based}\\
\hline
Deformable-DETR\cite{zhu2020deformable} & ICLR'20 & 40M & 14.2 & 25.7 & 13.8 & 8.0 & 21.7 & 30.1 & 19.2 & 44.8 & 13.7 & 6.3 & 15.4 & 24.9 & 34.2 \\
Conditional-DETR\cite{meng2021conditional} & ICCV'21 & 46M & 26.4 &  37.7 & 27.4 & 16.2 & 36.2 & 35.6 & 25.7 & 52.8 & 15.0 & 7.9 & 20.3 & 28.0 & 36.5 \\
DAB-DETR\cite{liu2021dab} & ICLR'22 & 55M & 27.8 & 40.8 & 26.9 & 16.4 & 35.5 & 39.8 & 27.2 & 55.1 & 20.6 & 10.3 & 22.5 & 31.9 & 37.2 \\
DINO\cite{zhang2022dino} & ICLR'23 & 56M & 26.8 & 44.2 & 28.9 & 17.5 & 37.3 & 41.3 & 28.9 & 59.4 & 22.4 & 12.5 & 22.7 & 34.7 & 42.8 \\
Co-DINO\cite{zong2023detrs} & ICCV'23 & 66M & 28.5 & \underline{46.7} & 29.9 &20.5 & \underline{38.2} & 46.3 & \underline{32.2} & \underline{61.1} & 28.9 & \underline{15.3} & 28.4 & \underline{38.9} & 48.4 \\
Salience-DETR\cite{hou2024salience} & CVPR'24 & 56M & \underline{28.9} & 46.5 & \underline{30.1} & \underline{21.2} & 38.0 & \underline{46.5} & 31.9 & 60.7 & \underline{29.1} & 15.0 & \underline{28.5} & 38.6 & \underline{48.5} \\
\rowcolor{gray!20}
NRQO & Ours & 63M & \textbf{32.1} & \textbf{51.7} & \textbf{33.9} & \textbf{24.5} & \textbf{41.9} & \textbf{53.5} & \textbf{33.4} & \textbf{62.8} & \textbf{30.1} & \textbf{16.2} & \textbf{29.8} & \textbf{39.6} & \textbf{49.7} \bigstrut[t]
\\
\bottomrule
\end{tabular}%
}
\label{Table_1}
\end{table*}

\vspace{-0.05in}
\subsection{Pairwise-Similarity Region Proposal Network}
Accurately matching anchors to ground truth boxes is crucial for generating high-quality positive queries for small objects. Traditional overlap-based methods, such as IoU, struggle with small objects, while distance-based methods like DotD~\cite{xu2021dot} and NWD~\cite{wang2021normalized} focus only on positions while neglecting shape, thereby reducing the quality of positive queries. To address these limitations, we propose the Pairwise-Similarity Region Proposal Network (PS-RPN), which integrates both position and shape similarity to generate a sufficient number of high-quality positive queries, significantly improving small object detection performance.

Specifically, PS-RPN introduces a novel Pairwise-Similarity (PS) metric that combines both position and shape similarities to more effectively quantify the alignment between anchors and ground truth boxes, thereby improving small object detection performance. The PS metric is defined as follows:
\begin{equation}
\label{11}
\mathcal{PS} = \exp \left(-(\mathcal{PS}_{position} + \mathcal{PS}_{shape})\right).
\end{equation}

Position similarity measures the distance between the center points of the ground truth and anchor, normalized by the dimensions of the bounding boxes,
\begin{equation}
\label{12}
\mathcal{PS}_{position} = \sqrt{\left(\frac{m \cdot (x_g - x) }{w_g + w}\right)^2 + \left(\frac{n \cdot (y_g - y) }{h_g + h}\right)^2}.
\end{equation}

Shape similarity captures the relative differences in width and height between the ground truth and anchor,
\begin{equation}
\label{13}
\mathcal{PS}_{shape} = \sqrt{\left(\frac{m \cdot (w_g - w)}{w_g + w}\right)^2 + \left(\frac{n \cdot (h_g - h)}{h_g + h}\right)^2}.
\end{equation}

Here, \( (x_g, y_g) \) and \( (x, y) \) denote the center coordinates of the ground truth and anchor, while \( w_g, w, h_g, h \) represent their widths and heights. The parameters \( m \) and \( n \) are defined as:
\begin{equation}
\label{14}
m = \frac{\sum_{i=1}^{M} \sum_{j=1}^{N_i} \sum_{k=1}^{Q_i} \frac{|x_{ij} - x_{ik}|}{(w_{ij} + w_{ik})}}{\sum_{i=1}^{M} N_i \cdot Q_i},
\end{equation}
\begin{equation}
\label{15}
n = \frac{\sum_{i=1}^{M} \sum_{j=1}^{N_i} \sum_{k=1}^{Q_i} \frac{|y_{ij} - y_{ik}|}{(h_{ij} + h_{ik})}}{\sum_{i=1}^{M} N_i \cdot Q_i},
\end{equation}
where $M$ is the number of training images, $N_i$ is the number of ground truth boxes in the $i$-th image, and $Q_i$ is the number of anchors in the same image. The variables $x_{ij}$ and $x_{ik}$ represent the $x$-coordinates of the $j$-th ground truth and the $k$-th anchor, respectively, with similar definitions applying to the $y$-coordinates.

Additionally, we adopt a Pairwise-Similarity label assignment strategy based on the traditional MaxIoUAssigner\cite{ren2015faster}, replacing IoU with PS. The positive, negative, and minimum positive thresholds are set to 0.7, 0.3, and 0.3, respectively. The training objective of PS-RPN is defined as follows:
\begin{equation}
\mathcal{L}_\text{PS-RPN} = \mathcal{L}_{\text{Cls}} + \mathcal{L}_{\text{Reg}},
\end{equation}
$\mathcal{L}_\text{Cls}$ and $\mathcal{L}_\text{Reg}$ denote classification and regression losses.

\subsection{Training and Optimization}
Based on the above analyses, the training objective of the proposed NRQO paradigm combines multiple objective functions from NT-FPN and PS-RPN,
\begin{equation}
\label{Eq_14} 
\mathcal{L}_\text{NRQO} =\alpha \cdot( \mathcal{L}_s + \mathcal{L}_{\hat{s}}) + \mathcal{L}_\text{PS-RPN},
\end{equation}
where $\alpha$ balances the contributions of different loss terms.

\section{Experiments}
\subsection{Dataset and Evaluation Metrics} 
We evaluate our NRQO paradigm on VisDrone 2019\cite{du2019visdrone}, SODA-D~\cite{cheng2023towards}, and COCO 2017~\cite{lin2014microsoft} datasets, using Average Precision (AP)\cite{lin2014microsoft} as the primary metric, along with $\text{AP}_{50}$ and $\text{AP}_{75}$ at IoU thresholds of 0.5 and 0.75. Objects in VisDrone 2019 and COCO 2017 are categorized by size into small (S), medium (M), and large (L), while SODA-D classifies objects into Small and Normal, further refining Small into extremely small (eS), relatively small (rS), and generally small (gS).

\subsection{Implementation Details and Baselines} 
All experiments are conducted on a single RTX 3090 GPU. The model was trained for 12 epochs using an AdamW optimizer with a batch size of 2, an initial learning rate of 2e-4, and a weight decay of 0.0001. For fairness, ResNet-50 is used as the backbone in all baseline models. We extensively compare our NRQO paradigm with various object detection baselines, including both \textbf{CNN-based} and \textbf{Transformer-based} methods.

\begin{table}[t]
\small
\centering
\renewcommand\tabcolsep{0.9pt}
\renewcommand\arraystretch{1.0}
\caption{Comparison with detection approaches on the COCO 2017.}
\vspace{-0.05in}
\begin{tabular}{l|c|ccc|ccc}
    \toprule
    \textbf{Methods}&\textbf{Ref} &\textbf{ $\text{AP}$}&\textbf{$\text{AP}_{50}$} & \textbf{$\text{AP}_{75}$}& \textbf{$\text{AP}_{S}$}& \textbf{$\text{AP}_{M}$}&\textbf{ $\text{AP}_{L}$}\\
    \hline
    \textbf{CNN-based} \\
    \hline
    Faster RCNN~\cite{ren2015faster}&NIPS'15&35.7&56.1&38.0&19.2&40.9&48.7\\
    FCOS~\cite{tian2019fcos}&ICCV'19&38.6&57.2&41.7&23.5&42.8&48.9\\
    ATSS~\cite{zhang2020bridging}&CVPR'20&39.3 & 57.5 & 42.7 & 22.9 & 42.9 & 51.2\\
    YOLOX~\cite{ge2021yolox}&CVPR'21&40.5 & 59.3 & 43.7 & 23.2 & 44.8 & 54.1\\
    Sparse RCNN~\cite{sun2021sparse}&CVPR'21&40.1 & 59.4 & 43.5 & 22.9 & 43.6 & 52.9\\
    \hline
    \textbf{Transformer-based} \\
    \hline
    Deformable-DETR~\cite{zhu2020deformable}& ICLR'20&31.8 & 51.4 & 33.5 & 15.0 & 35.7 & 44.7\\ 
    Conditional-DETR~\cite{meng2021conditional}& ICCV'21&32.2 & 52.1 & 33.4 & 13.9 & 34.5 & 48.7\\ 
    DAB-DETR~\cite{liu2021dab}& ICLR'22&38.0 & 60.3 & 39.8 & 19.2 & 40.9 & 55.4\\ 
    DINO~\cite{zhang2022dino}& ICLR'23&49.0 & 66.6 & 53.5 & 32.0 & 52.3 & 63.0\\ 
    Co-DINO~\cite{zong2023detrs}&ICCV'23&\underline{50.8} & \underline{68.5} & \underline{55.7} & \underline{33.9} & \underline{54.0} & \underline{64.8}\\
    Salience-DETR~\cite{hou2024salience} &CVPR'24& 49.2 & 67.1 & 53.8 & 32.7 & 53.0 & 63.1\\
    \rowcolor{gray!20}
    NRQO&Ours&\textbf{53.2}&\textbf{70.3}&\textbf{58.4}&\textbf{36.6}&\textbf{55.8}&\textbf{66.5} \bigstrut[t]\\
    \bottomrule
\end{tabular}
\label{Table_2}
\end{table}

\begin{figure*}[t]
\centerline{\includegraphics[width=\linewidth]{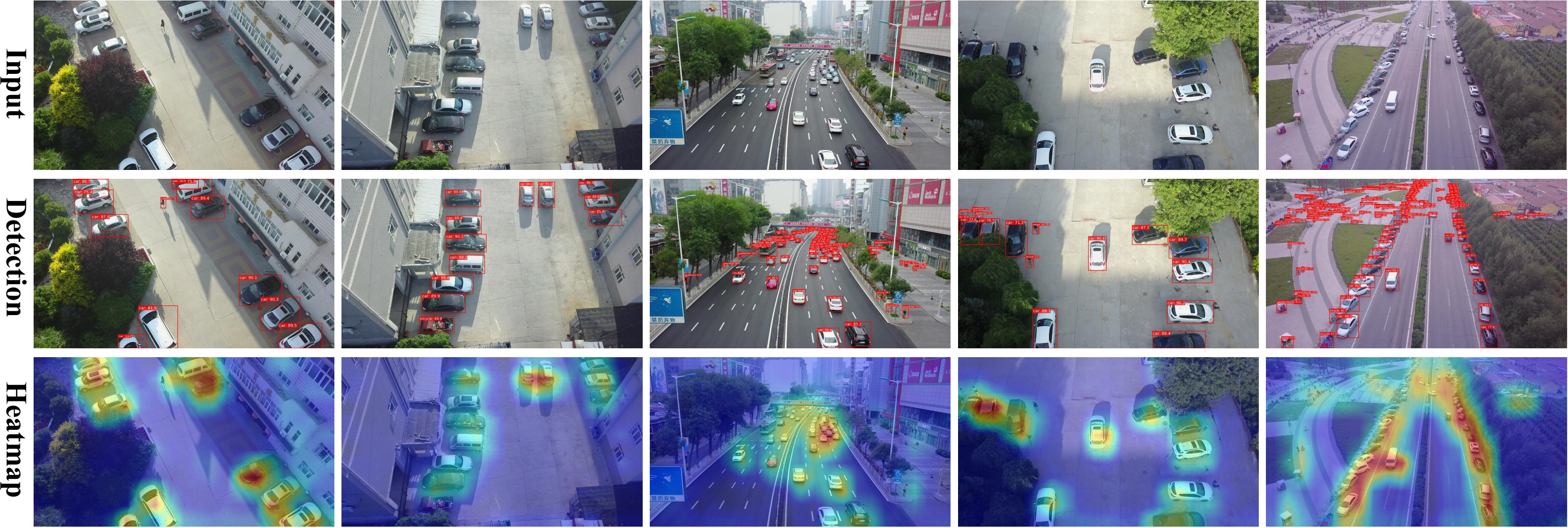}}
\caption{Visualization of detection results and query heatmaps for our proposed NRQO paradigm on the VisDrone 2019 dataset.}
\label{Fig_4}
\vspace{-0.1in}
\end{figure*}

\subsection{Comparisons with State-of-The-Arts} 
\subsubsection{Evaluation on VisDrone 2019} We first evaluate the performance of the proposed NRQO paradigm on VisDrone 2019, where it outperforms all baseline methods. As shown in Table~\ref{Table_1}, NRQO achieves a 6.1\% improvement in AP over leading CNN-based models such as CFInet (32.1\% vs. 26.0\%) and a 3.2\% improvement over state-of-the-art Transformer-based models like Salience-DETR (32.1\% vs. 28.9\%), demonstrating its effectiveness in detecting small objects.

\subsubsection{Evaluation on SODA-D} 
As shown in Table~\ref{Table_1}, NRQO achieves a 1.2\%  improvement in AP over the best-performing Transformer-based model, Co-DINO (33.4\% vs. 32.2\%). Notably, NRQO excels in detecting extremely small (AP$_{eS}$) and relatively small objects (AP$_{rS}$), with improvements of 0.9\% and 1.3\%, respectively. These results highlight its precision and robustness in handling small objects.

\subsubsection{Evaluation on COCO 2017} 
As shown in Table~\ref{Table_2}, NRQO achieves superior performance on the COCO 2017 dataset, with an AP of 53.2\%. It outperforms CNN-based models like YOLOX by 12.7\% and Transformer-based models like Co-DINO by 2.4\% (53.2\% vs. 50.8\%). This advantage is particularly evident in small (AP$_{S}$) and medium (AP$_{M}$) object categories, with gains of 2.7\% and 1.8\%, respectively.

\begin{table}[t]
\centering
\small
\renewcommand\tabcolsep{2pt}
\renewcommand\arraystretch{1.0}
\caption{Ablation studies on two modules.}
\vspace{-0.05in}
\begin{tabular}{c|cc|ccc|ccc}
         \toprule
          \textbf{Settings}&\textbf{NT-FPN}&\textbf{PS-RPN}&\textbf{$\text{AP}$} &\textbf{$\text{AP}_{50}$} & \textbf{$\text{AP}_{75}$}& \textbf{$\text{AP}_{S}$}& \textbf{$\text{AP}_{M}$}& \textbf{$\text{AP}_{L}$}\\
          \midrule
          I&\ding{55}&\ding{55} &28.5 &46.7&29.9&20.5&38.2&46.3   \\
          II&\ding{51} &\ding{55}&31.6&51.2&32.8&23.7&41.3&52.8 \\
          III&\ding{55}&\ding{51}&30.9&50.6&33.2&23.1&40.8&53.0\\ 
          IV&\ding{51}&\ding{51}&\textbf{32.1}&\textbf{51.7}&\textbf{33.9}&\textbf{24.5}&\textbf{41.9}&\textbf{53.5}\\
          \bottomrule
     \end{tabular}
\label{Table_3}
\vspace{-0.1in}
\end{table}

\subsection{Ablation Studies}
In the ablation studies, we systematically evaluate the impact of each component of our proposed NRQO paradigm on small object detection performance. As shown in Table~\ref{Table_3}, the results for “NRQO w/o NT-FPN” highlight the crucial role of NT-FPN in reducing noise during multi-scale feature fusion and enhancing small object representation, thereby improving the model’s ability to detect small objects. The results for “NRQO w/o PS-RPN” demonstrate that PS-RPN significantly enhances detection accuracy by incorporating both position and shape similarities, ensuring more precise anchor-to-ground-truth matching and generating higher-quality positive queries for small objects.

\subsection{Parameter Analysis} 
As shown in Table~\ref{Table_4}, the coefficient $\alpha$ in Eq. \ref{Eq_14} balances the influence of spatial and semantic losses. Both insufficient and excessive weighting of these losses negatively impact performance; however, adjusting the coefficient $\alpha$ demonstrates low sensitivity to overall performance. Consequently, we set $\alpha$ to 0.1 to achieve optimal precision.

\begin{table}[t]
\centering
\small
\renewcommand\tabcolsep{6.0pt}
\renewcommand\arraystretch{1.0}
\caption{Ablation on the coefficient of the training loss.}
\vspace{-0.05in}
\begin{tabular}{c|ccc|ccc}
         \toprule
          \textbf{$\alpha$}&\textbf{$\text{AP}$} &\textbf{$\text{AP}_{50}$} & \textbf{$\text{AP}_{75}$}& \textbf{$\text{AP}_{S}$}& \textbf{$\text{AP}_{M}$}& \textbf{$\text{AP}_{L}$}\\
          \midrule
          0.05&31.3&49.3&33.2&23.7&41.6&53.0   \\
          0.07&31.7&50.8&\textbf{34.0}&23.4&\textbf{42.2}&53.2 \\
          \textbf{0.1}&\textbf{32.1}&\textbf{51.7}&33.9&\textbf{24.5}&41.9&\textbf{53.5}\\ 
          0.5&31.4&50.5&33.6&21.8&41.3&53.1\\
          0.7&31.3&50.1&32.6&21.2&40.7&52.8\\
          \bottomrule
     \end{tabular}
\label{Table_4}
\vspace{-0.1in}
\end{table}

\subsection{Visualization Results} 
To further evaluate the performance of our proposed NRQO paradigm, we visualize detection results and query heatmaps on the VisDrone 2019 dataset. As shown in Fig.~\ref{Fig_4}, NRQO accurately detects and localizes small objects, with the detected objects clearly highlighted by red bounding boxes. The corresponding heatmaps demonstrate NRQO’s ability to capture essential features, highlighting both spatial and semantic details. These maps further showcase how NRQO maintains a strong focus on small objects, ensuring high-quality query generation, and effectively enhancing small object detection.

\section{Conclusions}
In this paper, we propose a novel Noise-Resilient Query Optimization (NRQO) paradigm to enhance small object detection. By seamlessly integrating the Noise-Tolerance Feature Pyramid Network (NT-FPN) and the Pairwise-Similarity Region Proposal Network (PS-RPN), NRQO effectively mitigates the challenges of inherent noise sensitivity and diminished query quality common in existing FPNs and label assignment strategies. Extensive experiments on three public datasets demonstrate the effectiveness of NRQO, achieving state-of-the-art performance. Our proposed NRQO paradigm shows significant promise for improving small object detection, making it highly applicable to real-world scenarios such as autonomous driving, surveillance, and medical imaging.


\bibliographystyle{IEEEtran}
\bibliography{icme2025references}
\end{document}